\pdfoutput=1

\documentclass[11pt]{article}

\usepackage[]{acl}

\usepackage{times}
\usepackage{latexsym}

\usepackage[T1]{fontenc}

\usepackage[utf8]{inputenc}

\usepackage{microtype}

\usepackage{soul}
\usepackage{tablefootnote}
\usepackage[normalem]{ulem}
\usepackage{multirow}
\usepackage{makecell}
\usepackage{enumitem}
\usepackage{adjustbox}
\usepackage{amsmath}
\usepackage{array}
\usepackage{graphicx}
\usepackage{comment}
\usepackage{booktabs}
\usepackage{xcolor,colortbl} 
\usepackage{algorithm}
\usepackage{bbm}
\usepackage{tabularx}
\usepackage{hyperref}

\usepackage{graphicx}
\usepackage{subcaption}
\usepackage{arydshln}
\usepackage{array}
\usepackage{booktabs}
\usepackage{pifont}

\definecolor{green}{rgb}{0.1,0.1,0.1}
\usepackage{tabulary}
\usepackage{blindtext}
\usepackage[noend]{algpseudocode}
\usepackage{stmaryrd}
\usepackage{tikz}

\definecolor{gitred}{HTML}{FDB8C0}
\definecolor{gitgreen}{HTML}{006400}
\definecolor{chocolate}{HTML}{D2691E}
\definecolor{maroon}{HTML}{800000}
\definecolor{indigo}{HTML}{4B0082}
\definecolor{green}{HTML}{008000}
\definecolor{orange}{HTML}{fc8d62}
\definecolor{purple}{HTML}{8da0cb}

\newcommand{\magdata}{{AcademicMKP}}
\newcommand{\ecomdata}{{EcommerceMKP}}

\title{Retrieval-Augmented Multilingual Keyphrase Generation with Retriever-Generator Iterative Training\thanks{~~Qingyu Yin and Zheng Li are corresponding authors. This work was mainly done while Rui Meng was a Ph.D. student at University of Pittsburgh.}}

\author{
 \makecell{Yifan Gao$^1$, Qingyu Yin$^{1\ddag}$, Zheng Li$^{1\ddag}$, Rui Meng$^2$, \\ Tong Zhao$^1$,  Bing Yin$^1$, Irwin King$^3$, Michael R. Lyu$^3$} \\
 {$^1$Amazon Inc.}  \quad {$^2$ Salesforce Research} \quad {$^3$ Chinese University of Hong Kong} \\
 \tt{$^1$\{yifangao,qingyy,amzzhe\}@amazon.com \quad $^2$ruimeng@salesforce.com} \\
}

\begin{document}
\maketitle
\begin{abstract}
Keyphrase generation is the task of automatically predicting keyphrases given a piece of long text.
Despite its recent flourishing, keyphrase generation on non-English languages haven't been vastly investigated.
In this paper, we call attention to a new setting named multilingual keyphrase generation and we contribute two new datasets, EcommerceMKP and AcademicMKP, covering six languages.
Technically, we propose a retrieval-augmented method for multilingual keyphrase generation to mitigate the data shortage problem in non-English languages.
The retrieval-augmented model leverages keyphrase annotations in English datasets to facilitate generating keyphrases in low-resource languages.
Given a non-English passage, a cross-lingual dense passage retrieval module finds relevant English passages.
Then the associated English keyphrases serve as external knowledge for keyphrase generation in the current language. 
Moreover, we develop a retriever-generator iterative training algorithm to mine pseudo parallel passage pairs to strengthen the cross-lingual passage retriever.
Comprehensive experiments and ablations show that the proposed approach outperforms all baselines.\footnote{The datasets are released at \url{https://github.com/Yifan-Gao/multilingual_keyphrase_generation}.} 

\end{abstract}






\section{Introduction}

Keyphrases are single or multi-word lexical units that best summarize a piece of text. As such, they are of great importance for indexing, categorizing, and mining in many information retrieval and natural language processing tasks~\cite{Jones1999PhrasierAS,Frank1999DomainSK,hulth-megyesi-2006-study,Dave2003MiningTP}.
Keyphrase generation is the task of automatically predicting keyphrases given a piece of long text.
Existing works on keyphrase generation mostly focus on English datasets \cite{gallina-etal-2019-kptimes,meng-etal-2017-deep} while keyphrase generation for languages other than English is still under-explored. 
Since search engines usually provide services to customers using different languages, multilingual keyphrase generation becomes a significant problem while it is still unknown how well existing keyphrase generation approaches perform in non-English languages. 
Nevertheless, there are two challenges we will face regarding multilingual keyphrase generation.
First, to the best of our knowledge, there is no large-scale dataset publicly available for training and benchmarking multilingual keyphrase generation models.
Building keyphrase datasets at a sufficient scale is difficult and costly.
Second, compared with the existing datasets in English, which can contain millions of data examples and cover a wide diversity of topics, the data resources in non-English languages are inherently scarce.
For example, in the domain of e-commerce, marketplaces using English have abundant customer queries to be used for keyphrase mining, while queries in some languages are relatively less than in English, which is probably because of a smaller size of user population or a shorter operation time. 

We start tackling these challenges by contributing two new datasets for multilingual keyphrase, which cover six languages and two domains.
The first dataset \ecomdata\ is collected from a real-world major e-commerce website. 
The product descriptions are used as the source text while the target keyphrases are collected from user search queries.
This dataset contains a total of 73k data examples, covering four different languages (Spanish, German, Italian and French).
The second multilingual keyphrase dataset \magdata\ lies in the academic domain, in which titles and abstracts are used as the source text and the author-provided keyphrases are deemed targets.
A total of 2,693 academic papers in Chinese and Korean are included in \magdata. 

To overcome the resource scarcity challenge in training multilingual models, we propose a retrieval-based method to leverage the keyphrase knowledge in large-scale English datasets. By investigating multilingual keyphrase data, we observe that data in different languages may talk about similar topics.
Therefore, we conjecture that passage-keyphrases pairs in English can be of help as an external knowledge base for multilingual keyphrase generation. To be specific,
given a passage in low-resource language \texttt{XX}, we propose to use a retrieval model to find multiple top-related passages in English.
These retrieved English passages provide high-quality English keyphrases that can be used as hints for generating keyphrases in other languages.
After that, the generator takes the code-mixed inputs, including the passage in language \texttt{XX} and retrieved English keyphrases, and predicts keyphrases in language \texttt{XX}.

In the cross-lingual retrieval training, parallel {passage-keyphrases} pairs between English and other languages are extremely limited. For example, in the e-commerce domain, only a small fraction of products have both English and non-English descriptions (being sold in multiple countries). 
{Such a data scarcity issue weakens the ability of cross-lingual knowledge acquisition from high-resource English keyphrases as intermediary, and finally hinders the potential of the retrieval-augmented keyphrase generation}.
To mitigate the problem, we propose a retriever-generator iterative training (RGIT) algorithm to automatically mine pseudo training parallel pairs from unlabeled data.
{Concretely, the retriever can dynamically adjust in terms of the current variations of generation performance between the proposed retrieval-augmented generator and the base one without the aid of retrieved English keyphrases.}
Starting from {insufficient seed parallel pairs, if the retrieved pseudo passage-keyphrases pairs in the current iteration can bring in higher generation results as the generator's feedback, those pseudo parallel data will be regarded as high quality and incorporated into the seed ones to further boost the retriever. Such cycle providing positive effects can be repeated until the increasing generation performance stopped.}

{We conduct}
extensive experiments on \ecomdata\ and \magdata\ {and demonstrate} that large-scale English datasets do provide useful knowledge for multilingual keyphrase generation.
The proposed retrieval-augmented method outperforms traditional extraction-based models, sequence-to-sequence neural models, and its variants.
Moreover, the RGIT algorithm boosts the retrieval performance significantly by mining over 20k pseudo-parallel passage pairs.
We also conduct detailed analyses to investigate the effectiveness of retriever-generator iterative training.

\begin{figure*}[t!]
\centering
\includegraphics[width=1.0\textwidth]{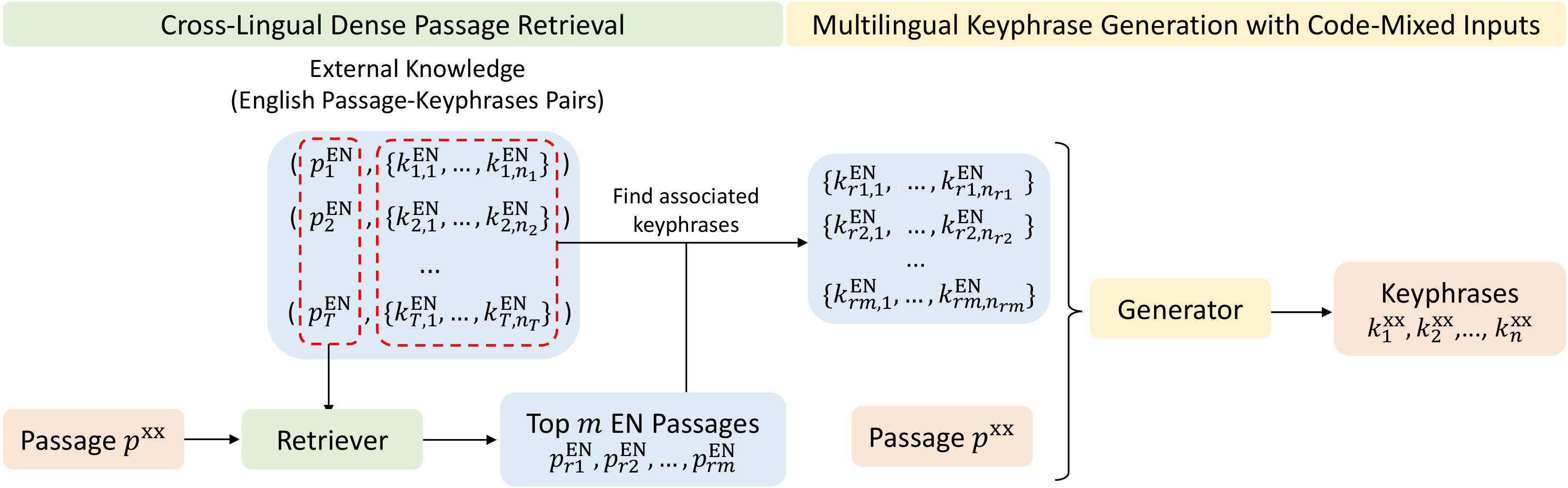}
\caption{Overview of our Retrieval-Augmented Multilingual Keyphrase Generation (RAMKG) framework. 
$p^{\texttt{XX}}, k^{\texttt{XX}}_i$ denote a passage and keyphrases in language \texttt{XX} (\texttt{XX} $\in$ \{ \texttt{DE}, \texttt{ES}, \texttt{FR}, \texttt{IT}, \texttt{KO}, \texttt{ZH} \}). 
$p^{\texttt{EN}}, k^{\texttt{EN}}_i$ denote relevant passages and keyphrases retrieved from the English dataset.
}
\label{fig:rag}
\end{figure*}

\section{Related Work}


\paragraph{Keyphrase Generation.}
The advance of neural language generation enables models to freely generate keyphrases according to the phrase importance and semantics, rather than extracting a list of sub-strings from the text \cite{witten1999KEA,liu2011gap,wang2016ptr}.
\citet{meng-etal-2017-deep} propose the first keyphrase generation model CopyRNN, which not only generates words based on a vocabulary but also points to words in the source text --- overcoming the barrier of predicting absent keyphrases.
Following this idea,
\citet{chen2018kp_correlation,zhao2019incorporating,ahmad-etal-2021-select} leverage the attention mechanism to reduce duplication and improve coverage.
\citet{Chen2019TitleGE,ye-wang-2018-semi,wang2019topic,liang-etal-2021-unsupervised} propose to leverage extra structure information (e.g., title, topic) to guide the generation.
\citet{chan-etal-2019-neural,luo-etal-2021-keyphrase-generation} propose a model using reinforcement learning, and \citet{swaminathan2020preliminary} propose using GAN for KPG.
\citet{chen-etal-2020-exclusive} introduce hierarchical decoding and exclusion mechanism to prevent models from generating duplicate phrases.
\citet{ye-etal-2021-one2set} propose to dynamically align target phrases to eliminate the influence of order, as highlighted by \citet{meng-etal-2021-empirical}.
\citet{mu2020keyphrase,liu2020reinforced,park-caragea-2020-scientific} use pre-trained language models for better representations of documents.

\paragraph{Retrieval Augmented Text Generation}
(RAG) recently shows great power in knowledge-intensive NLP tasks such as open-domain question answering, fact checking and entity linking \cite{NEURIPS2020_6b493230,petroni-etal-2021-kilt,guu2020realm}.
In RAG, a retriever (either sparse \cite{lee-etal-2019-latent} or dense \cite{karpukhin-etal-2020-dense}) searches for useful non-parametric knowledge from a knowledge base, then a generator combines the non-parametric retrieved knowledge with its parametric knowledge, learned during pre-training, for solving the task.
Different from these tasks, keyphrase generation is not a knowledge-intensive task but we treat the English passage-keyphrase training data as our knowledge.
Similar approaches have been investigated in neural machine translation \cite{Gu2018SearchEG,cai-etal-2021-neural}, dialogue \cite{weston-etal-2018-retrieve}, and knowledge-base QA \cite{das-etal-2021-case}.
In keyphrase generation, \citet{chen-etal-2019-integrated,ye2021heterogeneous,kim-etal-2021-structure} retrieve similar documents from training data to produce more accurate keyphrases. However, their retrieval module is a non-parametric model and cannot be generalized in the multilingual setting due to the large vocabulary gap between languages. 

\section{Task Definition}
In this paper, we aim to tackle the keyphrase generation task in a multilingual setting, which means one model of desire is capable of generating keyphrases in any language that it has been trained with.
The benefits of having a single keyphrase generation model for multiple languages are threefold:
(1) Collecting keyphrase annotation for individual language can be prohibitively expensive;
(2) Training and deploying separate models for each language is laborious; 
(3) Joint training of multiple languages can alleviate the resource scarcity by utilizing rich monolingual data. 

Formally, we define the multilingual keyphrase generation task as follows.
Given a piece of text $p^{\texttt{XX}}$ in language \texttt{XX}, our goal is to predict its corresponding keyphrases $k^{\texttt{XX}}_{1}, k^{\texttt{XX}}_{2}, ..., k^{\texttt{XX}}_{n}$ in language \texttt{XX}, where $n$ is the total number of target keyphrases for this text $p^{\texttt{XX}}$.
In this study, \texttt{XX} can be German (\texttt{DE}), Spanish (\texttt{ES}), Italian (\texttt{IT}), French (\texttt{FR}), Korean (\texttt{KO}) or Chinese (\texttt{ZH}).

\begin{figure*}[t!]
\centering
\includegraphics[width=1.0\textwidth]{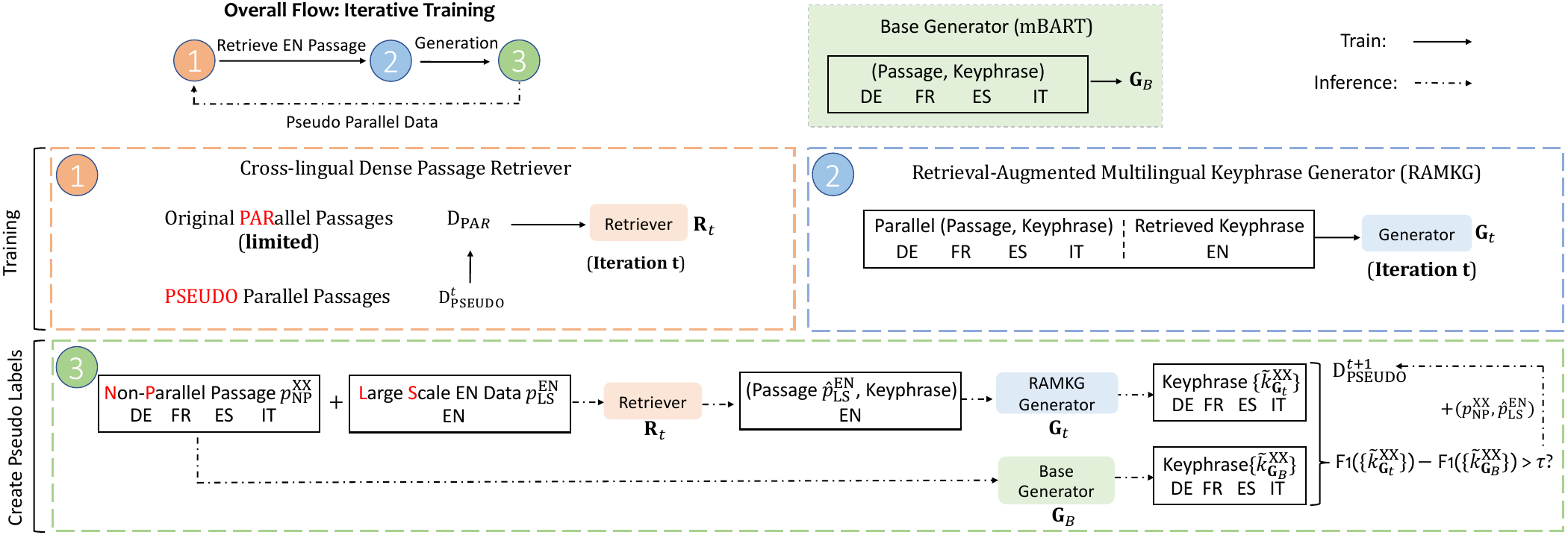}
\caption{Retriever-Generator Iterative Training for Parallel Passage Mining
}
\label{fig:selftraining}
\end{figure*}

\section{Model} \label{sec:framework}


Scarcity of resources is one of the topmost challenges for multilingual tasks, which is also the case for multilingual keyphrase generation. One may find it difficult to collect enough text data in languages other than English, much less the annotation of keyphrases in specific domains.
To overcome this problem, we propose a retrieval-augmented approach to make use of the relatively rich resources in English.
The motivation for our proposed retrieval-augmented approach comes from an observation from data: texts and keyphrases expressed in different languages usually share common topics or knowledge concepts.
For example, in e-commerce websites, it is often the case that the same products are sold in different marketplaces/countries.
Thus these products as well as their keyphrases, though exhibited in different languages, can share a high semantic similarity.
In other words, given a text in language $p^\texttt{AA}$, if we could find a similar text in language $p^\texttt{BB}$, its associated keyphrases $k^{\texttt{BB}}$ may serve as a good hint for the to-be-generated keyphrases $k^{\texttt{AA}}$ in language \texttt{AA}.
Since English has the most abundant text-keyphrases pairs in both e-commerce and academic papers domains, its resource can be treated as a non-parametric keyphrase knowledge base, which provides texts in English covering a wide range of topics and concepts, as well as the associated high-quality keyphrases.

As shown in Fig.~\ref{fig:rag}, our framework consists of a retrieval step and a generation step:
\begin{enumerate}[leftmargin = 15pt,topsep=0pt,noitemsep]
    \item \textit{Retrieval Step}: 
    given a source passage $p^{\texttt{XX}}$ in language \texttt{XX}, the cross-lingual retriever first finds $m$ semantically relevant English passages $p^{\texttt{EN}}_{1}, p^{\texttt{EN}}_{2}, ..., p^{\texttt{EN}}_{m}$. 
    Each retrieved English passage $p^{\texttt{EN}}_{j}$ has its associated $n_j$ English keyphrases $k^{\texttt{EN}}_{j,1}$, $k^{\texttt{EN}}_{j,2}$, ..., $k^{\texttt{EN}}_{j,n_j}$. 
    These retrieved English keyphrases are taken as external knowledge for keyphrase generation in step 2. 
    \item \textit{Generation Step}: 
    taking the source text $p^{\texttt{XX}}$ in language \texttt{XX} and all retrieved English keyphrases \{$k^{\texttt{EN}}_{1,1}$, ..., $k^{\texttt{EN}}_{1,n_1}$\}, ..., \{$k^{\texttt{EN}}_{m,1}$, ..., $k^{\texttt{EN}}_{m,n_m}$\} as inputs, the generation module concatenates them as a sequence and generates keyphrases in target language \texttt{XX}.
\end{enumerate}


\subsection{Cross-Lingual Dense Passage Retrieval} \label{sec:retriever}

The cross-lingual retriever includes a passage encoder $E_P(\cdot)$ and a query encoder $E_Q(\cdot)$.
The passage encoder $E_P(\cdot)$ maps millions of English passages into $d$-dimensional vectors and builds indices for all English passages using FAISS~\cite{Johnson2021BillionSS} offline.
At inference time, the passage in language \texttt{XX}
goes through the query encoder $E_Q(\cdot)$ and is converted into a $d$-dimensional vector.
Then the cross-lingual retriever performs a KNN search to retrieve $m$ English passages whose vectors are closest to the query vector measured by the dot product similarity: $\text{sim} ( p^{\texttt{XX}}, p^{\texttt{EN}_j} ) = {E_Q(p^{\texttt{XX}})}^{\top} E_P(p^{\texttt{EN}}_j)$.

\paragraph{Passage Encoder.} 
Since naive lexical similarity can hardly handle text matching across languages, we resort to a BERT-based dense passage retriever~\cite{karpukhin-etal-2020-dense}, expecting the contextualized semantic matching can retrieve similar passages accurately and robustly.
In order to meet the demand of multilingual representation, we utilize multilingual pre-trained model mBERT~\cite{devlin-etal-2019-bert} to encode passages into 768-dimensional vectors. 

\paragraph{Training.}
Since the output vectors of mBERT are not aligned across languages, we need extra alignment training to ensure that similar passages in different languages can be mapped into near regions in the high-dimensional space.
Given a passage $p^{\texttt{XX}}_i$ in language \texttt{XX}, we take its corresponding English passage $p^{\texttt{EN}+}_i$ as the positive example and randomly select $n$ negative passages $p^{\texttt{EN}-}_{i,1}, ..., p^{\texttt{EN}-}_{i,n}$ in the English corpus.
The dense retriever is trained by optimizing the negative log likelihood loss of the positive English passage.

In the e-commerce domain, we select the positive passage according to product metadata.
For a product sold in both \texttt{EN} and \texttt{XX} marketplaces, we regard its bilingual product descriptions $(p^{\texttt{XX}}_i, p^{\texttt{EN}+}_i)$ as a parallel passage pair, i.e., positive training example. 
For the domain of academic paper, we notice that papers with parallel text is very rare.
Therefore, we develop an automatic approach to mine parallel abstract pairs of English and the target language.
Specifically, we adopt an off-the-shelf bi-text mining tool named Sentence Transformers~\cite{reimers-gurevych-2019-sentence} to mine pseudo parallel pairs.
Given two datasets in different languages, we encode passages using LaBSE~\cite{Feng2020LanguageAB}, the current best method for learning language-agnostic sentence embeddings for 109 languages, and then parallel passages can be extracted through nearest-neighbor retrieval and filtered by setting a fixed threshold over a margin-based similarity score, as proposed in~\cite{artetxe-schwenk-2019-margin}.

\subsection{Multilingual Keyphrase Generation with Code-Mixed Inputs} \label{sec:generator}

Given the top $m$ retrieved English passages $p^\texttt{EN}_1$, ..., $p^\texttt{EN}_m$, we find their associated keyphrases in the dataset: $\{ k^{\texttt{EN}}_{1,1}$, ..., $k^{\texttt{EN}}_{1,n_1}\}$, ..., $\{k^{\texttt{EN}}_{m,1}$, ..., $k^{\texttt{EN}}_{m,n_m}\}$.
We utilize mBART~\cite{liu-etal-2020-multilingual-denoising}, a multilingual denoising pre-trained sequence-to-sequence language model, to integrate information from multiple languages. 
Different from machine translation which maps a sentence a the source language to a target language, our multilingual keyphrase generation model takes code-mixed inputs -- a combination of retrieved English keyphrases
$\{ k^{\texttt{EN}}_{1,1}$, ..., $k^{\texttt{EN}}_{1,n_1}\}$, ..., $\{k^{\texttt{EN}}_{m,1}$, ..., $k^{\texttt{EN}}_{m,n_m}\}$ from $m$ retrieved English passages and the source passage $p^{\texttt{XX}}$ in the target language \texttt{XX}.

We concatenate retrieved English keyphrases with a delimiter token \texttt{[SEP]}, and add special tokens to separate different inputs: \texttt{[ENKPS]} for retrieved keyphrases and \texttt{[CTX]} for the source passage.
Besides, we follow the fine-tuning setup of mBART by adding the language identifier \texttt{[XX]} (e.g. \texttt{[DE]} for German) at the end of the input sequence to denote the current input language:\\
\begin{footnotesize}
\texttt{[ENKPS]} $k^{\texttt{EN}}_{1,1}$ \texttt{[SEP]} ...  \texttt{[SEP]} $k^{\texttt{EN}}_{m,n_m}$ \texttt{[CTX]} $p^\texttt{XX}$ \texttt{[XX]}.
\end{footnotesize}
The training target is a sequence of concatenated keyphrases $k^{\texttt{XX}}_{1}$, ..., $k^{\texttt{XX}}_{n}$, separated by a special token \texttt{[SEP]}. \texttt{[XX]}, the language identifier of the current language, is also added at the beginning of the target sequence to indicate the target language: 
\begin{center}
    \texttt{[XX]} $k^{\texttt{XX}}_{1}$ \texttt{[SEP]} $k^{\texttt{XX}}_{2}$ \texttt{[SEP]} ... \texttt{[SEP]} $k^{\texttt{XX}}_{n}$.
\end{center}


\begin{algorithm}[!t]
\fontsize{8pt}{10pt}\selectfont
\caption{
    Parallel Passage Mining via Iterative Training
}\label{alg:pseudo}
\begin{algorithmic}[1]
\State \textbf{Input}: 
(1) Parallel data $D_{\texttt{PAR}}=\{(p^{\texttt{EN}}_\texttt{PAR}, p^{\texttt{XX}}_\texttt{PAR}, k^{\texttt{EN}}_\texttt{PAR}, k^{\texttt{XX}}_\texttt{PAR})\}$, 
(2) Non-parallel data $D_{\texttt{NP}}=\{(p^{\texttt{XX}}_\texttt{NP}, k^{\texttt{XX}}_\texttt{NP})\}$,
(3) Large-scale English corpus $D_{\texttt{LS}}=\{(p^\texttt{EN}_\texttt{LS}, k^{\texttt{EN}}_\texttt{LS})\}$.
\State \textbf{Output}: Pseudo parallel passage pairs $D_\texttt{PSEUDO}$

\State {\protect\color{gitgreen}  {\textbf{$\rhd$ 0. Train Seq2Seq baseline w/o retrieved keyphrases}}}
\State $\mathbf{G}_B \gets train( \{(p^{\texttt{XX}}_\texttt{PAR}, k^{\texttt{XX}}_\texttt{PAR})\} \in D_{\texttt{PAR}})$ 
\State $D_{\texttt{PSEUDO}}^{0} \gets \{\}$ ~~ {\protect\color{gitgreen}  {\textit{// Pseudo parallel passage pairs  }}}

\For{$t \in \{0...T-1 \}$}
    \State {\protect\color{gitgreen}  {\textbf{$\rhd$ 1. Train retriever on pseudo and parallel data }}}
    \State $\mathbf{R}_t \gets train(\{(p^{\texttt{EN}}_\texttt{PAR}, p^{\texttt{XX}}_\texttt{PAR})\} \in D_{\texttt{PSEUDO}}^{t} \cup D_{\texttt{PAR}})$ 

    \State {\protect\color{gitgreen}  {\textbf{$\rhd$ 2. Train retrieval-augmented generator on $D_\texttt{PAR}$ }}}
    \For{each $(p^{\texttt{XX}}_\texttt{PAR}, k^{\texttt{XX}}_{\texttt{PAR}}) \in D_\texttt{PAR}$}
        \State $\hat{p}^{\texttt{EN}}_{\texttt{LS}} \gets \mathbf{R}_t(p^{\texttt{XX}}_\texttt{PAR}, D_{\texttt{LS}})$  ~~ {\protect\color{gitgreen}  {\textit{// Retrieve \texttt{EN} passages}}}
        \State $\{\hat{k}^{\texttt{EN}}_{\texttt{LS}}\} \gets \hat{p}^{\texttt{EN}}_{\texttt{LS}}$ ~~ {\protect\color{gitgreen}  {\textit{// Find associated \texttt{EN} keyphrases}}}
    \EndFor
    \State {\protect\color{gitgreen}  {\textit{// Train retrieval-augmented generator with \texttt{EN} keyphrases}}}
    \State $\mathbf{G}_t \gets train(\{(p^{\texttt{XX}}_\texttt{PAR}, k^{\texttt{XX}}_\texttt{PAR})\} \in D_{\texttt{PAR}}, \{\hat{k}^{\texttt{EN}}_\texttt{LS}\})$ 
    
    \State {\protect\color{gitgreen}  {\textbf{$\rhd$ 3. Create pseudo parallel passage pairs }}}
    \State $D_{\texttt{PSEUDO}}^{t+1} \gets \{\}$
    \For{each $p^{\texttt{XX}}_\texttt{NP} \in D_{\texttt{NP}}$}
        \State $\hat{p}^{\texttt{EN}}_{\texttt{LS}} \gets \mathbf{R}_t(p^{\texttt{XX}}_\texttt{NP}, D_{\texttt{LS}})$  ~~ {\protect\color{gitgreen}  {\textit{// Retrieve \texttt{EN} passages }}}
        \State $\{\hat{k}^{\texttt{EN}}_{\texttt{LS}}\} \gets \hat{p}^{\texttt{EN}}_{\texttt{LS}}$ ~~ {\protect\color{gitgreen}  {\textit{// Find associated \texttt{EN} keyphrases}}}
        
        \State {\protect\color{gitgreen}  {\textit{// Predict keyphrases w/o and w/ \texttt{EN} keyphrases}}}
        \State $\{\tilde{k}^{\texttt{XX}}_{\texttt{NP},\mathbf{G}_B}\} \gets \mathbf{G}_B(p^{\texttt{XX}}_\texttt{NP})$
        \State $\{\tilde{k}^{\texttt{XX}}_{\texttt{NP},\mathbf{G}_t}\} \gets \mathbf{G}_t(p^{\texttt{XX}}_\texttt{NP}, \{\hat{k}^{\texttt{EN}}_\texttt{LS}\})$
        
        \State {\protect\color{gitgreen}  {\textit{// If adding \texttt{EN} keyphrases leads to better keyphrase predictions, the retrieved  \texttt{EN} passages are taken as positive examples}}}
        
        \If{$\text{F1}(\{\tilde{k}^{\texttt{XX}}_{\texttt{NP},\mathbf{G}_t}\}) - \text{F1}(\{\tilde{k}^{\texttt{XX}}_{\texttt{NP},\mathbf{G}_B}\}) > \tau$}
            \State $D_{\texttt{PSEUDO}}^{t+1} \gets D_{\texttt{PSEUDO}}^{t+1} \cup \{(\hat{p}^{\texttt{EN}}_\texttt{NP}, p^{\texttt{XX}}_\texttt{NP})\}$
        \EndIf
    \EndFor

\EndFor
\State \Return $D_{\texttt{PSEUDO}}^{T}$
\end{algorithmic}
\end{algorithm}

\subsection{\mbox{Retriever-Generator Iterative Training}} \label{sec:selftrain}
In spite of having utilized parallel passage pairs to align the multilingual representations of the retrieval module, it remains a concern because the parallel passage pairs between English and non-English languages account for only a small portion of the whole multilingual dataset. 
For example, in a popular e-commerce platform, only a small percentage of products (less than 10\%) have both English and non-English descriptions.
Without enough quality parallel pairs, the cross-lingual dense passage retriever may not work well to find relevant English passages.
Consequently, associated English keyphrases may provide little help for multilingual keyphrase generation.

To make the multilingual keyphrase generation generalize better to any target languages or domains without reliance on numerous parallel passage pairs, we propose an iterative training method to mine parallel passages which requires only a small number of initial parallel pairs of bootstrap the process.
Since our ultimate goal of the retriever model is to provide useful external knowledge for multilingual keyphrase generation, we mine parallel passage pairs (English and a non-English language) according to whether the retrieved English passage-keyphrases pairs could help the keyphrase generation for the target non-English language \texttt{XX}.
For example, let $p^\texttt{EN}_a$ and $p^\texttt{EN}_b$ be two retrieved English passages for a passage $p^\texttt{XX}$ in target language, if the associated keyphrases of $p^\texttt{EN}_a$ provide more useful information for generating the keyphrases of  $p^\texttt{XX}$ than $p^\texttt{EN}_b$, then $(p^\texttt{EN}_a, p^\texttt{XX})$ would be considered as a better parallel passage pair.
That said, we expect the mined pseudo parallel passage pairs to be of high quality according to the retrieval score, at the same time they can be directly helpful for training the generation module. 

The proposed iterative training approach is sketched in Algo.~\ref{alg:pseudo} and Fig.~\ref{fig:selftraining}.
Given a \underline{L}arge-\underline{S}cale keyphrase dataset in English $D_{\texttt{LS}}=\{(p^\texttt{EN}_\texttt{LS}, k^{\texttt{EN}}_\texttt{LS})\}$ and a smaller one $D_{\texttt{XX}}=\{(p^\texttt{XX}, k^{\texttt{XX}})\}$ in target language $\texttt{XX}$, we denote the set of annotated parallel examples (bilingual passages in English and other languages) as \textbf{\underline{PAR}allel split} $D_\texttt{PAR} = \{ (p^{\texttt{EN}}_\texttt{PAR}, p^{\texttt{XX}}_\texttt{PAR}, k^{\texttt{EN}}_\texttt{PAR}, k^{\texttt{XX}}_\texttt{PAR}) \}$, in which $\{(p^{\texttt{XX}}_\texttt{PAR}, k^{\texttt{XX}}_\texttt{PAR})\}$ comes from the $\texttt{XX}$ dataset while $\{(p^{\texttt{EN}}_\texttt{PAR}, k^{\texttt{EN}}_\texttt{PAR})\}$ comes from the English dataset.
The remaining data examples in the target dataset $D_{\texttt{XX}}$ have no annotated corresponding English examples in $D_{\texttt{LS}}$ (the pairs may exist but are not known yet), and we name this set as the \textbf{\underline{N}on-\underline{P}arallel split} $D_{\texttt{NP}}=\{(p^{\texttt{XX}}_\texttt{NP}, k^{\texttt{XX}}_\texttt{NP})\}$.
We firstly fine-tune a mBART using only keyphrases data of target language $\{(p^{\texttt{XX}}_\texttt{PAR}, k^{\texttt{XX}}_\texttt{PAR})\}$ in $D_\texttt{PAR}$ (Line 4).
Then we start a loop to mine pseudo parallel passage pairs for refining the passage retriever.
Each iteration is expected to bring in a higher quality of pseudo passage pairs, resulting in a better performance of retriever, with three steps:
\begin{enumerate}[leftmargin = 15pt,topsep=0pt,noitemsep]
    \item We train a retriever $\mathbf{R}_t$ using existing available \texttt{EN}-\texttt{XX} passage pairs $\{(p^{\texttt{EN}}_\texttt{PAR}, p^{\texttt{XX}}_\texttt{PAR})\}$ from both parallel passage data $D_{\texttt{PAR}}$ and most up-to-date pseudo passage data $D_{\texttt{PSEUDO}}^{t}$ (Line 8).
    \item We train a retrieval-augmented model $\mathbf{G}_t$ using multilingual passages $\{(p^{\texttt{XX}}_\texttt{PAR}, k^{\texttt{XX}}_\texttt{PAR})\}$ from the parallel data $D_{\texttt{PAR}}$ and retrieved English keyphrases $\{\hat{k}^{\texttt{EN}}_\texttt{LS}\}$ from the English dataset $D_{\texttt{LS}}$ (Line 14). To get the retrieved English keyphrases for each passage $p^{\texttt{XX}}_\texttt{PAR}$, we take the retriever $\mathbf{R}_t$ trained in step (1) to do a KNN search for passages $\hat{p}^{\texttt{EN}}_{\texttt{LS}}$ in \texttt{EN} $D_{\texttt{LS}}$ and find their associated keyphrases $\{\hat{k}^{\texttt{EN}}_\texttt{LS}\}$ (Line 10-12).
    \item For each passage $p^{\texttt{XX}}_\texttt{NP}$ in the non-aligned dataset $D_{\texttt{NP}}$, we also retrieve English passages $\hat{p}^{\texttt{EN}}_{\texttt{LS}}$ and keyphrases $\{\hat{k}^{\texttt{EN}}_{\texttt{LS}}\}$ from $D_{\texttt{LS}}$ (Line 18-19).
    Then the retrieved English passage $\hat{p}^{\texttt{EN}}_{\texttt{LS}}$ will be taken as the parallel text to $p^{\texttt{XX}}_\texttt{NP}$ if its associated keyphrases $\{\hat{k}^{\texttt{EN}}_{\texttt{LS}}\}$ provide \textit{useful} information.
    The \textit{usefulness} is measured by the keyphrase generation performance (F-score) between the retrieval-augmented generation model $\mathbf{G}_t$ and the base model $\mathbf{G}_0$ that does not use \texttt{EN} keyphrases (Line 21-25).
\end{enumerate}

After $T$ iterations, we train the retriever on the pseudo data $D_\texttt{PSEUDO}^T$ and fine-tune it on the parallel data $D_\texttt{PAR}$.
Then we treat it as our final retriever and train the generation model in Sec.~\ref{sec:generator}.

\begin{table}[!t]
  \centering
  \resizebox{1.0\columnwidth}{!}{
    \begin{tabular}{c|ccccccc}
    \toprule
    \multirow{0.5}{*}{Language} & \multicolumn{1}{c}{\shortstack{Train \\ Size}} & \multicolumn{1}{c}{\shortstack{Dev \\ Size}} & \multicolumn{1}{c}{\shortstack{Test \\ Size}} & \multicolumn{1}{c}{\shortstack{Passage Length \\ (Avg/Std/Mid)}} & \multicolumn{1}{c}{\shortstack{\#Keyphrases \\ (Avg/Std/Mid)}} &  \multicolumn{1}{c}{\shortstack{Absent \\ Kps\%}} \\
    \midrule
    \multicolumn{7}{c}{\magdata\ Dataset} \\
    \midrule
    Chinese (ZH) & 1,110  & 158   & 319   & 217/48/207 & 5/1/5 & 27.2\% \\
    Korean (KO) & 774   & 110   & 222   & 115/31/111 & 4/1/4 & 37.7\% \\
    Total & 1,884  & 268   & 541   & 171/57/155 & 4/1/4 & 31.3\% \\
    \midrule
    \multicolumn{7}{c}{\ecomdata\ Dataset} \\
    \midrule
    German (DE) & 23,997 & 1,411  & 2,825  & 157/79/141 & 10/5/8 & 57.1\% \\
    Spanish (ES) & 12,222 & 718   & 1,440  & 159/84/139 & 9/5/7 & 54.6\% \\
    French (FR) & 16,986 & 998   & 2,000  & 163/84/144 & 9/5/8 & 63.0\% \\
    Italian (IT) & 9,163  & 538   & 1,081  & 167/84/152 & 8/3/7 & 42.6\% \\
    Total & 62,368 & 3,665  & 7,346  & 161/82/143 & 9/5/7  & 56.4\% \\
    \bottomrule
    \end{tabular}%
    }
  \caption{\magdata\ \& \ecomdata\ Dataset}
  \label{tab:dataset-stat}%
\end{table}%

\section{Datasets}\label{sec:dataset}


\paragraph{\ecomdata\ Dataset} is collected from a popular E-commerce shopping platform.
There are four languages we consider for building \ecomdata: German (DE), French (FR), Spanish (ES) and Italian (IT).
The title, product description, and bullet description provided by manufacturers are concatenated and treated as source input.
The keyphrases of each product are selected from search queries under the following protocol.
First, given a product, we only keep search queries that lead to purchases and treat them as effective queries.
Then phrases are chunked from these effective queries using AutoPhrase~\cite{shang2018automated} and further ranked by their frequency.
Our assumption is: the more times a phrase appears in effective search queries of a product, the more important a phrase is.
Finally a threshold is set to filter out unimportant phrases.
Under this protocol, we receive 73k examples over four languages. The statistics are shown in Table~\ref{tab:dataset-stat}.

We collect the passages and keyphrases under the same protocol for the English (EN) dataset and name it as \ecomdata-EN.
In total the English dataset contains 3 million passage-keyphrases pairs.
To obtain the parallel passage pairs for training the cross-lingual dense passage retriever, we pair the product descriptions in different languages according to the product identification information.
We select a total of 1,247 parallel passages from \ecomdata\ training set which include 480 passages for \texttt{DE}-\texttt{EN}, 244 for \texttt{ES}-\texttt{EN}, 340 for \texttt{FR}-\texttt{EN}, and 183 for \texttt{IT}-\texttt{EN}.
Besides, we keep 1,000 parallel passage pairs in the DEV set of \ecomdata\ to evaluate the performance of retrieval and bi-text mining.

\begin{table*}[!ht]
  \centering
  \resizebox{1.0\textwidth}{!}{
    \begin{tabular}{l|ccccccccccccccc}
    \toprule
    \multirow{2}[2]{*}{Model} & \multicolumn{3}{c}{German (DE)} & \multicolumn{3}{c}{Spanish (ES)} & \multicolumn{3}{c}{French (FR)} & \multicolumn{3}{c}{Italian (IT)} & \multicolumn{3}{c}{Average} \\
          & P     & R     & F1    & P     & R     & F1    & P     & R     & F1    & P     & R     & F1    & P     & R     & F1 \\
    \midrule 
    \multicolumn{16}{c}{{{\texttt{Unsupervised Statistical Keyphrase Extraction}}}} \\  
    \midrule 
    KP-Miner & 9.13  & 2.35  & 3.34  & 14.56 & 4.59  & 6.22  & 7.78  & 2.76  & 3.62  & 22.51 & 7.71  & 10.31 & 11.80 & 3.69  & 5.01 \\
    YAKE  & 2.31  & 26.54 & 4.17  & 3.26  & 36.86 & 5.89  & 2.47  & 27.29 & 4.43  & 3.87  & 50.12 & 7.10  & 2.77  & 32.24 & 5.01 \\
    \midrule 
    \multicolumn{16}{c}{{{\texttt{Unsupervised Graph-based Keyphrase Extraction}}}} \\  
    \midrule 
    TextRank & 5.77  & 9.00  & 6.39  & 7.15  & 11.59 & 7.97  & 5.45  & 8.33  & 5.94  & 8.18  & 15.27 & 9.85  & 6.31  & 10.25 & 7.09 \\
    TopicalPageRank & 3.59  & 14.10 & 5.34  & 5.07  & 22.08 & 7.74  & 3.75  & 15.49 & 5.65  & 5.22  & 25.43 & 8.26  & 4.16  & 17.71 & 6.33 \\
    PositionRank & 5.24  & 11.67 & 6.90  & 7.86  & 19.36 & 10.75 & 5.56  & 13.50 & 7.56  & 7.98  & 21.46 & 11.34 & 6.24  & 15.12 & 8.49 \\
    MultipartiteRank & 5.09  & 8.76  & 6.15  & 7.56  & 14.25 & 9.46  & 5.31  & 9.86  & 6.59  & 7.72  & 15.93 & 10.12 & 6.02  & 11.19 & 7.50 \\
    \midrule 
    \multicolumn{16}{c}{{{\texttt{Supervised Feature-based Keyphrase Extraction}}}}\\
    \midrule 
    Kea   & 9.64  & 17.73 & 11.92 & 12.69 & 24.64 & 16.11 & 8.81  & 16.76 & 11.05 & 14.16 & 30.18 & 18.81 & 10.68 & 20.65 & 13.52 \\
    \midrule 
    \multicolumn{16}{c}{{{\texttt{Neural-based Supervised Keyphrase Generation}}}} \\
    \midrule 
    CopyRNN & 13.48 & 7.59  & 9.08  & 21.11 & 12.40 & 14.78 & 14.79 & 8.48  & 10.04 & 34.66 & 20.07 & 24.28 & 18.45 & 10.61 & 12.69 \\
    Transformer & 29.92 & 25.40 & 26.22 & 34.01 & 30.57 & 30.83 & 29.17 & 24.03 & 25.18 & 44.05 & 43.34 & 42.56 & 32.60 & 28.68 & 29.25 \\
    mBART (monolingual) & 44.91 & 39.59 & 40.04 & 47.70 & 44.79 & 44.17 & 42.64 & 36.81 & 37.72 & 57.98 & 58.16 & 56.25 & 46.76 & 42.58 & 42.60 \\
    mBART (multilingual) & 45.78 & 40.93 & 41.09 & 48.43 & 44.99 & 44.57 & 43.21 & 38.40 & 38.72 & 60.37 & 58.91 & 57.87 & 47.75 & 43.68 & 43.60 \\
    mBART + EN Joint Train & 45.91 & 39.90 & 40.64 & 49.27 & 43.92 & 44.52 & 43.21 & 37.48 & 38.26 & 59.21 & 57.03 & 56.42 & 47.79 & 42.55 & 43.08 \\
    mBART + EN Pretrain & 45.77 & 40.76 & 41.05 & 48.34 & 44.94 & 44.58 & 42.96 & 38.10 & 38.45 & 60.24 & 58.51 & 57.59 & 47.64 & 43.46 & 43.47 \\ 
    \midrule 
    RAMKG (\textbf{Ours}) & 46.88 & 41.90 & 42.14 & 49.35 & 45.89 & 45.49 & 43.79 & 39.48 & 39.61 & 60.89 & 59.51 & 58.43 & 48.59 & 44.61 & 44.50 \\
    RAMKG + RGIT (\textbf{Ours}) & \textbf{48.11} & \textbf{43.05} & \textbf{43.30} & \textbf{50.54} & \textbf{47.04} & \textbf{46.64} & \textbf{45.07} & \textbf{40.77} & \textbf{40.86} & \textbf{62.35} & \textbf{60.75} & \textbf{59.87} & \textbf{49.86} & \textbf{45.81} & \textbf{45.73} \\
    \bottomrule
    \end{tabular}%
    }
  \caption{Main results on the EcommerceMKP dataset. The best results are in bold. (RGIT: Iterative-training)}
  \label{tab:main-amaz}%
  \vspace{-1em}
\end{table*}%

\begin{table}[!t]
  \centering
  \fontsize{9.5}{10.2}\selectfont
  \resizebox{1.0\columnwidth}{!}{
    \begin{tabular}{l|ccccccccc}
    \toprule
    \multirow{2}[2]{*}{Model} & \multicolumn{3}{c}{Chinese (ZH)} & \multicolumn{3}{c}{Korean (KO)} & \multicolumn{3}{c}{Average} \\
          & P     & R     & F1    & P     & R     & F1    & P     & R     & F1 \\
    \midrule
    mBART (mono.) & 32.52 & 31.50 & 31.50 & 24.57 & 26.46 & 24.96 & 29.26 & 29.43 & 28.81 \\
    mBART (multi.) & 32.48 & 32.27 & 31.85 & 27.03 & 26.93 & 26.44 & 30.25 & 30.08 & 29.63 \\
    mBART + Joint & 31.10 & 30.38 & 30.23 & 27.36 & 26.85 & 26.58 & 29.57 & 28.93 & 28.73 \\
    mBART + Pretrain &   32.72    &   29.77    &   30.66    &   27.56    &   25.30    &    27.56   &   30.60    &   27.94    & 28.68  \\
    RAMKG (\textbf{Ours}) & 33.40 & 32.66 & 32.45 & 28.30 & \textbf{28.17} & 27.68 & 31.31 & 30.82 & 30.49 \\
    RAMKG + RGIT (\textbf{Ours}) & \textbf{34.38} & \textbf{33.05} & \textbf{33.15} & \textbf{29.33} & 27.87 & \textbf{28.00} & \textbf{32.31} & \textbf{30.92} & \textbf{31.04} \\
    \bottomrule
    \end{tabular}%
    }
  \caption{Results on AcademicMKP (mono: monolingual, multi: multilingual, Joint: EN Joint Train, Pretrain: EN Pretrain, RGIT: Iterative-training).}
  \label{tab:main-pub}%
\end{table}%

\paragraph{\magdata\ Dataset} is collected from the academic domain.
We take the title and abstract of each paper as the source text and author-provided keywords as the target output.
All papers are sampled from Microsoft Academic Graph~\cite{Sinha2015AnOO}, a web-scale academic entity graph that contains multiple types of scholarly entities and relationships: field of study, author, institution, abstract, venue, and keywords.
We use Spacy (\url{https://spacy.io/}) to detect the language of abstracts and keyphrases, and choose two languages Chinese (ZH) and Korean (KO) to construct the \magdata\ dataset.
Since Microsoft Academic Graph (MAG) is automatically crawled and constructed, we find some of its data is extremely noisy. For example, keyphrases might be missing or contain incorrect information such as titles, author names, and publication venues. Some of the abstracts are incomplete.
Therefore, we hire three annotators to manually examine the samples from MAG dataset. 
Data examples with incomplete abstracts are removed. 
We further manually verify the metadata of all examples and correct their keyphrase information if needed.
Finally, 2,693 high-quality data examples of scientific papers in the computer science domain are collected to constitute \magdata.

Besides the multilingual \magdata\ dataset, we use KP20K~\cite{meng-etal-2017-deep} as the English data for retrieval-augmented generation.
KP20K has 560k abstract-keyphrases pairs collected from various online digital libraries in computer science domain.
The threshold is set as 1.03 for passage mining and we receive 841 parallel passage pairs from \magdata\ training set, in which 433 \texttt{ZH}-\texttt{EN} passage pairs and 384 for \texttt{KO}-\texttt{EN}.

\section{Experimental Setup}
\subsection{Evaluation Metrics}
\paragraph{Keyphrase Generation.} 
Let the ground truth keyphrases be $Y: k_1, k_2, ..., k_n$ and the predicted keyphrases be $\tilde{Y}: \tilde{k}_1, \tilde{k}_2, ..., \tilde{k}_M$, we compute the precision ($P@M$), recall ($R@M$) and F-score ($F_1@M$) between $Y$ and $\tilde{Y}$ as $P@M = \frac{|Y \cap \tilde{Y}|}{|\tilde{Y}|}, R@M = \frac{|Y \cap \tilde{Y}|}{{|Y|}}, F_1@M = \frac{2 \times P \times R}{P + R}$,
where $|Y|$ denotes the number of keyphrases in the gold set $Y$.
We only consider exact match of two keyphrases (with some post-processing such as lowercase) for $|Y \cap \tilde{Y}|$.
Then the average are computed for all languages in the test set.

\paragraph{Passage Retrieval.}
The quantity of retrieved English passages directly influences how much external knowledge could be utilized for keyphrase generation.
Therefore, we evaluate the top-k recall (k=1,2,5,10,20) on the DEV set for evaluating retrieval performance.

\subsection{Baselines and Ablations}
We consider following baselines and ablations: 
\textbf{1) Unsupervised Statistical Keyphrase Extraction}: {KP-Miner}~\cite{el-beltagy-rafea-2010-kp}, {YAKE}~\cite{Campos2020YAKEKE}; 
\textbf{2) Unsupervised Graph-based Keyphrase Extraction}: {TextRank}~\cite{mihalcea-tarau-2004-textrank}, {TopicalPageRank}~\cite{Sterckx2015TopicalWI}, {PositionRank}~\cite{florescu-caragea-2017-positionrank}, {MultipartiteRank}~\cite{boudin-2018-unsupervised};
\textbf{{3) Supervised Feature-based Keyphrase Extraction}}: {KEA}~\cite{Witten1999PracticalAK};
\textbf{4) Neural Supervised Keyphrase Generation}: {CopyRNN}, {Transformer};
\textbf{5) mBART (monolingual)}: separately trained 6 mBART models on each language;
\textbf{6) mBART (multilingual)}: a single mBART model on all languages;
\textbf{7) mBART + EN Joint Train}: a mBART model jointly trained on the multilingual data and English data (KP20K~\cite{meng-etal-2017-deep} for \magdata; \ecomdata-EN for \ecomdata).
\textbf{8) mBART + EN Pretrain}: a mBART firstly pre-trained on the English data and then fine-tuned on the multilingual data. 
\textbf{9) RAMKG (Ours)}: The Retrieval-Augmented Multilingual Keyphrase Generation model (Sec.~\ref{sec:retriever} \& \ref{sec:generator}).
\textbf{10) RAMKG + RGIT ({Ours})}: RAMKG improved with retriever-generator iterative training (RGIT) (Sec.~\ref{sec:selftrain}).

\section{Results and Analyses}\label{sec:results}

\subsection{Main Results}
Main results are shown in Table~\ref{tab:main-amaz}~\&~\ref{tab:main-pub} for \ecomdata\ and \magdata\ respectively, and we make the following observations:
\begin{itemize}[leftmargin = 15pt,topsep=0pt,noitemsep]
    \item The unsupervised approaches, both statistical-based and graph-based, have robust results across all languages. PositionRank performs the best among all unsupervised approaches.
    \item The supervised approaches consistently outperform unsupervised approaches. The feature-based approach KEA receives a high recall by predicting more keyphrases while the CopyRNN receives a high precision. 
    Different from the results on English keyphrase generation where Transformer and CopyRNN are comparable, the Transformer beats the CopyRNN by a large margin in the multilingual scenario.
    \item We observe that jointly training on all languages (mBART multilingual) receives better results than separately training on each language (mBART monolingual). This implies the ability of locating and summarizing key information is transferable across languages.
    \item Comparing different approaches using external large-scale English data, we find that our proposed RAMKG outperforms both ``EN Joint Train'' and ``EN Pretrain''. This is because the retrieval-augmented approach provides auxiliary knowledge information as part of the input to the generation module, while the other two variants have to ``infuse'' the knowledge learned from English data to model parameters. 
    Moreover, ``EN Joint Train'' and ``EN Pretrain'' have no positive effect on \magdata\ dataset (Table~\ref{tab:main-pub}). Compared with multilingual and English data are from the same website, there is still a domain gap between papers (multilingual) in \magdata\ and papers (English) in KP20K.
    \item The retriever-generator self-training (RAMKG + Iter) alleviates the data scarcity issue with the help of stronger retriever: since the retriever can find more relevant English keyphrases, it leads to a general improvement on keyphrase performance across languages. 
\end{itemize}

\subsection{\mbox{Effect of Iterative Training}} \label{sec:selftrain-analysis}

\paragraph{Retrieval Results}
We investigate the effect of retriever-generator iterative training by comparing the retrieval recall for models trained w/o and w/ the mined pseudo parallel passage pairs.
Results on the DEV set of \ecomdata\ are shown in the Table~\ref{tab:retrieval-result}. 
With additional mined pseudo parallel passage pairs, the retriever improves the Recall@5 from 50.1\% to 72.4\%.
And therefore, the better retrieved English keyphrases lead to a better generation performance (44.50 vs. 45.73 in Table~\ref{tab:main-amaz}).

\paragraph{Quantity and Quality of Pseudo Parallel Passages}
We show the quantity and quality of mined pseudo parallel pairs in Table~\ref{tab:label-acc}.
After each iteration of passage mining, our algorithm can consistently find around 20k passage pairs from \ecomdata\ training set, which are nearly 20 times of the initial data.
To assess the quality of mined passage pairs, we examine the label accuracy using the 1,000 parallel passage pairs from the DEV set of \ecomdata.
Results in Table~\ref{tab:label-acc} show that while the passage mining finds a similar number of pseudo passage pairs, the labelling accuracy does increase from 28.0\% to 47.0\%.
This is because the better pseudo parallel data improves the retriever, and the stronger retriever results in a better generator, which in turn leads to more relevant passages.

\begin{table}[!t]
  \centering
  \resizebox{0.8\columnwidth}{!}{
    \begin{tabular}{cccccc}
    \toprule
    Recall @ Top K  & 1     & 2     & 5     & 10    & 20 \\
    \midrule
    RAMKG & 26.4\% & 36.8\% & 50.1\% & 59.2\% & 67.3\% \\
    RAMKG + RGIT & 45.8\% & 58.3\% & 72.4\% & 79.5\% & 85.2\% \\
    \bottomrule
    \end{tabular}%
    }
  \caption{Retrieval recall on \ecomdata\ DEV set.}
  \label{tab:retrieval-result}%
\end{table}%

\begin{table}[!t]
  \centering
  \resizebox{1.0\columnwidth}{!}{
    \begin{tabular}{ccccccc}
    \toprule
    Iteration & 1     & 2     & 3     & 4     & 5     & 6 \\
    \midrule
    \# Pseudo Passages & 20,288 & 21,402 & 19,942 & 21,241 & 22,343 & 21,557 \\
    Label Accuracy \% & 28.0\% & 37.9\% & 40.5\% & 44.2\% & 45.1\% & 47.0\% \\
    \bottomrule
    \end{tabular}%
    }
  \caption{Number of pseudo parallel passages and their accuracy on \ecomdata\ DEV set in different iterations of parallel passages mining.}
  \label{tab:label-acc}%
\end{table}%

\begin{figure}[t!]
\centering
\includegraphics[width=1.0\columnwidth]{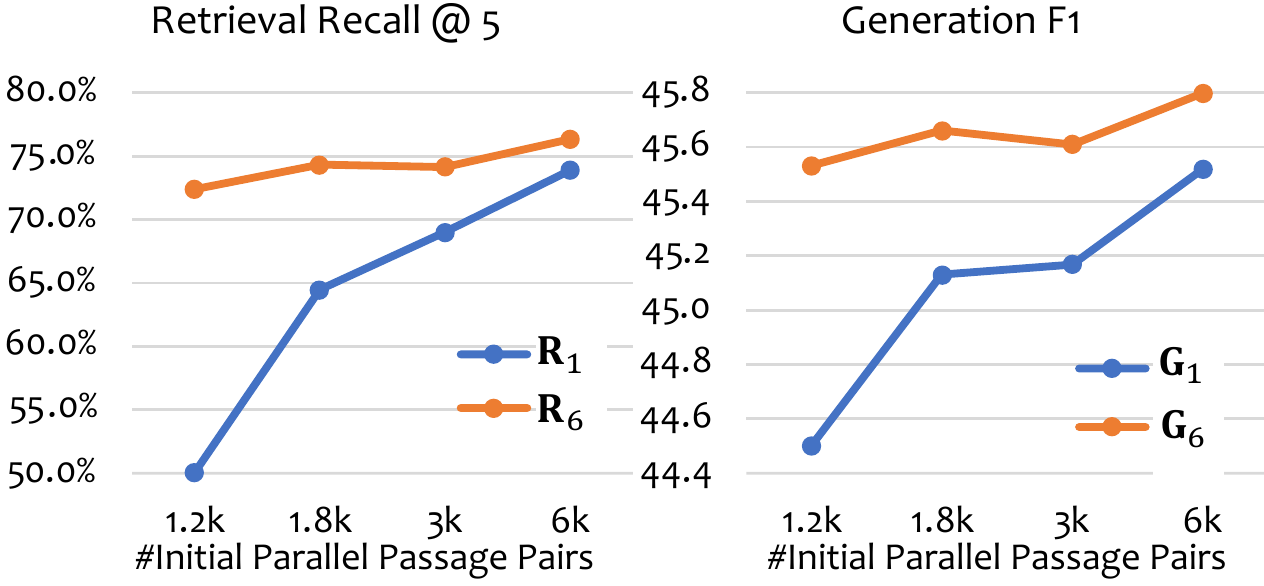}
\caption{Performance of passage retrieval and keyphrase generation on \ecomdata, with different number of initial parallel data for iterative training.
}
\label{fig:align-percent}
\vspace{-1em}
\end{figure}


\paragraph{Initializing with Different Amount of Parallel Data}

To investigate the impact of initial parallel passage on the training, we conduct experiments by varying the number of parallel passage pairs on \ecomdata, from 1.2k (default setting) to 6k instances. 
We compare the single-round training (i.e., training with initial data) and iterative training after six rounds (in which round we generally obtain the best retrieval recall), on both passage retrieval ($\mathbf{R}_1$ \& $\mathbf{R}_6$) and keyphrase generation ($\mathbf{G}_1$ \& $\mathbf{G}_6$).
Results are shown in Fig.~\ref{fig:align-percent}. 
We observe that (1) the score of iterative training consistently increase when more annotated parallel data is available; (2) our iterative training demonstrates great robustness with limited parallel data (e.g. 1.2k pairs), while the benefit gradually diminishes while more parallel data becomes available.



\begin{figure}[t!]
\small
\begin{tabular}{p{0.95\columnwidth}}
\hline\hline
\textbf{Product Description (German)}: \textbf{Steiff} 113437 \textbf{Soft Cuddly Friends} Honey Teddybär, \textbf{grau}, 38 cm. Bereits der Name des \textbf{Soft Cuddly Friends} Honey Teddybär sagt es schon aus: der 38 cm große Freund mit seinem honigsüßen Lächeln begeistert alle Kinderherzen ...\\
\rowcolor{lightgray}{{(Translation in English)}}: Steiff 113437 Soft Cuddly Friends Honey teddy bear, gray, 38 cm. The name of the Soft Cuddly Friends Honey Teddy bear already says it all: the 38 cm tall friend with his honey-sweet smile delights all children's hearts ... \\
\textbf{Gold Keyphrases (German)}: \textit{steiff kuscheltier}; \textit{steiff teddy}; \textbf{soft cuddly friend}; \textbf{steiff}; \textit{baer}; \textbf{grau}. \\
\rowcolor{lightgray}{{(Translation in English)}}: steiff cuddly toy; steiff teddy; soft cuddly friend; steiff; bear; grey. \\
\hline
\textbf{Retrieved English Keyphrases}: steiff teddy bear; teddy bear; my first; grey; honey; sweetheart; steiff bear; pink; vintage; steiff stuffed animal; steiff; terry; soft; jimmy. \\
\textbf{Predicted Keyphrases (German)}: \textcolor{green}{\textit{steiff kuscheltier}}; \textcolor{green}{\textit{steiff teddy}}; \textcolor{green}{\textbf{soft cuddly friend}}; \textcolor{green}{\textbf{steiff}}; \textcolor{green}{\textit{baer}}; \textcolor{green}{\textbf{grau}}; \textcolor{red}{\textit{jimmy}}. \\
\rowcolor{lightgray}{{(Translation in English)}}: steiff cuddly toy; steiff teddy; soft cuddly friend; steiff; bear; grey; jimmy. \\
\hline\hline                
\end{tabular}
\caption{Case study on the EcommerceMKP dataset. The present keyphrases (keyphrases shown in the description) are in \textbf{bold} while absent keyphrases are in \textit{italics}. 
Correct predictions are in \textcolor{green}{green} while wrong predictions are in \textcolor{red}{red}.
}
\label{fig:case}
\end{figure}
\subsection{Case Study}
Fig.~\ref{fig:case} exhibits an example of our model's prediction. Given a product description in German, the model retrieves several English keyphrases and generates keyphrases in German accurately (translations in English are also provided). 
Through this example, we find that the retrieved English keyphrases do provide certain useful information such as ``steiff teddy bear'', ``grey'' and ``soft'', while it also brings some noise such as ``my first'', ``sweetheart'' and ``vintage''.
Although there is a wrong prediction ``jimmy'' caused by the retrieved English keyphrases, the improvement in results shows that the benefits of retrieved knowledge outweigh the noise it introduces.
Moreover,the retrieved keyphrases are only regarded as a supplement to the original passage, and the generator can automatically focus on the informative parts from both inputs through self-attention.
Our retrieval-augmented multilingual keyphrase generation can tolerate some noise from the retrieved English keyphrases and predict better keyphrases based on these external knowledge.

\section{Conclusion}
In this study, we investigate a novel task setting -- multilingual keyphrase generation -- and 
contribute two new multilingual keyphrase generation datasets covering multiple domains and languages.
Furthermore, we propose a retrieval-augmented multilingual keyphrase generation framework with retriever-generator iterative training.
Results show that the proposed approach outperforms a wide range of baselines.



\bibliography{anthology,custom}

\begin{thebibliography}{56}
\expandafter\ifx\csname natexlab\endcsname\relax\def\natexlab#1{#1}\fi

\bibitem[{Ahmad et~al.(2021)Ahmad, Bai, Lee, and
  Chang}]{ahmad-etal-2021-select}
Wasi Ahmad, Xiao Bai, Soomin Lee, and Kai-Wei Chang. 2021.
\newblock \href {https://doi.org/10.18653/v1/2021.acl-long.111} {Select,
  extract and generate: Neural keyphrase generation with layer-wise coverage
  attention}.
\newblock In \emph{Proceedings of the 59th Annual Meeting of the Association
  for Computational Linguistics and the 11th International Joint Conference on
  Natural Language Processing (Volume 1: Long Papers)}, pages 1389--1404,
  Online. Association for Computational Linguistics.

\bibitem[{Artetxe and Schwenk(2019)}]{artetxe-schwenk-2019-margin}
Mikel Artetxe and Holger Schwenk. 2019.
\newblock \href {https://doi.org/10.18653/v1/P19-1309} {Margin-based parallel
  corpus mining with multilingual sentence embeddings}.
\newblock In \emph{Proceedings of the 57th Annual Meeting of the Association
  for Computational Linguistics}, pages 3197--3203, Florence, Italy.
  Association for Computational Linguistics.

\bibitem[{Boudin(2018)}]{boudin-2018-unsupervised}
Florian Boudin. 2018.
\newblock \href {https://doi.org/10.18653/v1/N18-2105} {Unsupervised keyphrase
  extraction with multipartite graphs}.
\newblock In \emph{Proceedings of the 2018 Conference of the North {A}merican
  Chapter of the Association for Computational Linguistics: Human Language
  Technologies, Volume 2 (Short Papers)}, pages 667--672, New Orleans,
  Louisiana. Association for Computational Linguistics.

\bibitem[{Cai et~al.(2021)Cai, Wang, Li, Lam, and Liu}]{cai-etal-2021-neural}
Deng Cai, Yan Wang, Huayang Li, Wai Lam, and Lemao Liu. 2021.
\newblock \href {https://doi.org/10.18653/v1/2021.acl-long.567} {Neural machine
  translation with monolingual translation memory}.
\newblock In \emph{Proceedings of the 59th Annual Meeting of the Association
  for Computational Linguistics and the 11th International Joint Conference on
  Natural Language Processing (Volume 1: Long Papers)}, pages 7307--7318,
  Online. Association for Computational Linguistics.

\bibitem[{Campos et~al.(2020)Campos, Mangaravite, Pasquali, Jorge, Nunes, and
  Jatowt}]{Campos2020YAKEKE}
Ricardo Campos, V{\'{\i}}tor Mangaravite, Arian Pasquali, Al{\'{\i}}pio Jorge,
  C{\'{e}}lia Nunes, and Adam Jatowt. 2020.
\newblock \href {https://doi.org/10.1016/j.ins.2019.09.013} {Yake! keyword
  extraction from single documents using multiple local features}.
\newblock \emph{Inf. Sci.}, 509:257--289.

\bibitem[{Chan et~al.(2019)Chan, Chen, Wang, and King}]{chan-etal-2019-neural}
Hou~Pong Chan, Wang Chen, Lu~Wang, and Irwin King. 2019.
\newblock \href {https://doi.org/10.18653/v1/P19-1208} {Neural keyphrase
  generation via reinforcement learning with adaptive rewards}.
\newblock In \emph{Proceedings of the 57th Annual Meeting of the Association
  for Computational Linguistics}, pages 2163--2174, Florence, Italy.
  Association for Computational Linguistics.

\bibitem[{Chen et~al.(2018)Chen, Zhang, Wu, Yan, and
  Li}]{chen2018kp_correlation}
Jun Chen, Xiaoming Zhang, Yu~Wu, Zhao Yan, and Zhoujun Li. 2018.
\newblock \href {https://doi.org/10.18653/v1/D18-1439} {Keyphrase generation
  with correlation constraints}.
\newblock In \emph{Proceedings of the 2018 Conference on Empirical Methods in
  Natural Language Processing}, pages 4057--4066, Brussels, Belgium.
  Association for Computational Linguistics.

\bibitem[{Chen et~al.(2019{\natexlab{a}})Chen, Chan, Li, Bing, and
  King}]{chen-etal-2019-integrated}
Wang Chen, Hou~Pong Chan, Piji Li, Lidong Bing, and Irwin King.
  2019{\natexlab{a}}.
\newblock \href {https://doi.org/10.18653/v1/N19-1292} {An integrated approach
  for keyphrase generation via exploring the power of retrieval and
  extraction}.
\newblock In \emph{Proceedings of the 2019 Conference of the North {A}merican
  Chapter of the Association for Computational Linguistics: Human Language
  Technologies, Volume 1 (Long and Short Papers)}, pages 2846--2856,
  Minneapolis, Minnesota. Association for Computational Linguistics.

\bibitem[{Chen et~al.(2020)Chen, Chan, Li, and King}]{chen-etal-2020-exclusive}
Wang Chen, Hou~Pong Chan, Piji Li, and Irwin King. 2020.
\newblock \href {https://doi.org/10.18653/v1/2020.acl-main.103} {Exclusive
  hierarchical decoding for deep keyphrase generation}.
\newblock In \emph{Proceedings of the 58th Annual Meeting of the Association
  for Computational Linguistics}, pages 1095--1105, Online. Association for
  Computational Linguistics.

\bibitem[{Chen et~al.(2019{\natexlab{b}})Chen, Gao, Zhang, King, and
  Lyu}]{Chen2019TitleGE}
Wang Chen, Yifan Gao, Jiani Zhang, Irwin King, and Michael~R. Lyu.
  2019{\natexlab{b}}.
\newblock \href {https://doi.org/10.1609/aaai.v33i01.33016268} {Title-guided
  encoding for keyphrase generation}.
\newblock In \emph{The Thirty-Third {AAAI} Conference on Artificial
  Intelligence, {AAAI} 2019, The Thirty-First Innovative Applications of
  Artificial Intelligence Conference, {IAAI} 2019, The Ninth {AAAI} Symposium
  on Educational Advances in Artificial Intelligence, {EAAI} 2019, Honolulu,
  Hawaii, USA, January 27 - February 1, 2019}, pages 6268--6275. {AAAI} Press.

\bibitem[{Das et~al.(2021)Das, Zaheer, Thai, Godbole, Perez, Lee, Tan,
  Polymenakos, and McCallum}]{das-etal-2021-case}
Rajarshi Das, Manzil Zaheer, Dung Thai, Ameya Godbole, Ethan Perez, Jay~Yoon
  Lee, Lizhen Tan, Lazaros Polymenakos, and Andrew McCallum. 2021.
\newblock \href {https://doi.org/10.18653/v1/2021.emnlp-main.755} {Case-based
  reasoning for natural language queries over knowledge bases}.
\newblock In \emph{Proceedings of the 2021 Conference on Empirical Methods in
  Natural Language Processing}, pages 9594--9611, Online and Punta Cana,
  Dominican Republic. Association for Computational Linguistics.

\bibitem[{Dave et~al.(2003)Dave, Lawrence, and Pennock}]{Dave2003MiningTP}
Kushal Dave, Steve Lawrence, and David~M. Pennock. 2003.
\newblock \href {https://doi.org/10.1145/775152.775226} {Mining the peanut
  gallery: opinion extraction and semantic classification of product reviews}.
\newblock In \emph{Proceedings of the Twelfth International World Wide Web
  Conference, {WWW} 2003, Budapest, Hungary, May 20-24, 2003}, pages 519--528.
  {ACM}.

\bibitem[{Devlin et~al.(2019)Devlin, Chang, Lee, and
  Toutanova}]{devlin-etal-2019-bert}
Jacob Devlin, Ming-Wei Chang, Kenton Lee, and Kristina Toutanova. 2019.
\newblock \href {https://doi.org/10.18653/v1/N19-1423} {{BERT}: Pre-training of
  deep bidirectional transformers for language understanding}.
\newblock In \emph{Proceedings of the 2019 Conference of the North {A}merican
  Chapter of the Association for Computational Linguistics: Human Language
  Technologies, Volume 1 (Long and Short Papers)}, pages 4171--4186,
  Minneapolis, Minnesota. Association for Computational Linguistics.

\bibitem[{El-Beltagy and Rafea(2010)}]{el-beltagy-rafea-2010-kp}
Samhaa~R. El-Beltagy and Ahmed Rafea. 2010.
\newblock \href {https://aclanthology.org/S10-1041} {{KP}-miner: Participation
  in {S}em{E}val-2}.
\newblock In \emph{Proceedings of the 5th International Workshop on Semantic
  Evaluation}, pages 190--193, Uppsala, Sweden. Association for Computational
  Linguistics.

\bibitem[{Feng et~al.(2020)Feng, Yang, Cer, Arivazhagan, and
  Wang}]{Feng2020LanguageAB}
Fangxiaoyu Feng, Yinfei Yang, Daniel Cer, Naveen Arivazhagan, and Wei Wang.
  2020.
\newblock \href {http://arxiv.org/abs/2007.01852} {Language-agnostic {BERT}
  sentence embedding}.
\newblock \emph{CoRR}, abs/2007.01852.

\bibitem[{Florescu and Caragea(2017)}]{florescu-caragea-2017-positionrank}
Corina Florescu and Cornelia Caragea. 2017.
\newblock \href {https://doi.org/10.18653/v1/P17-1102} {{P}osition{R}ank: An
  unsupervised approach to keyphrase extraction from scholarly documents}.
\newblock In \emph{Proceedings of the 55th Annual Meeting of the Association
  for Computational Linguistics (Volume 1: Long Papers)}, pages 1105--1115,
  Vancouver, Canada. Association for Computational Linguistics.

\bibitem[{Frank et~al.(1999)Frank, Paynter, Witten, Gutwin, and
  Nevill{-}Manning}]{Frank1999DomainSK}
Eibe Frank, Gordon~W. Paynter, Ian~H. Witten, Carl Gutwin, and Craig~G.
  Nevill{-}Manning. 1999.
\newblock \href {http://ijcai.org/Proceedings/99-2/Papers/002.pdf}
  {Domain-specific keyphrase extraction}.
\newblock In \emph{Proceedings of the Sixteenth International Joint Conference
  on Artificial Intelligence, {IJCAI} 99, Stockholm, Sweden, July 31 - August
  6, 1999. 2 Volumes, 1450 pages}, pages 668--673. Morgan Kaufmann.

\bibitem[{Gallina et~al.(2019)Gallina, Boudin, and
  Daille}]{gallina-etal-2019-kptimes}
Ygor Gallina, Florian Boudin, and Beatrice Daille. 2019.
\newblock \href {https://doi.org/10.18653/v1/W19-8617} {{KPT}imes: A
  large-scale dataset for keyphrase generation on news documents}.
\newblock In \emph{Proceedings of the 12th International Conference on Natural
  Language Generation}, pages 130--135, Tokyo, Japan. Association for
  Computational Linguistics.

\bibitem[{Gu et~al.(2018)Gu, Wang, Cho, and Li}]{Gu2018SearchEG}
Jiatao Gu, Yong Wang, Kyunghyun Cho, and Victor O.~K. Li. 2018.
\newblock Search engine guided neural machine translation.
\newblock In \emph{AAAI}.

\bibitem[{Guu et~al.(2020)Guu, Lee, Tung, Pasupat, and Chang}]{guu2020realm}
Kelvin Guu, Kenton Lee, Zora Tung, Panupong Pasupat, and Ming-Wei Chang. 2020.
\newblock {REALM}: Retrieval-augmented language model pre-training.
\newblock \emph{International Conference on Machine Learning (ICML)}.

\bibitem[{Hulth and Megyesi(2006)}]{hulth-megyesi-2006-study}
Anette Hulth and Be{\'a}ta~B. Megyesi. 2006.
\newblock \href {https://doi.org/10.3115/1220175.1220243} {A study on
  automatically extracted keywords in text categorization}.
\newblock In \emph{Proceedings of the 21st International Conference on
  Computational Linguistics and 44th Annual Meeting of the Association for
  Computational Linguistics}, pages 537--544, Sydney, Australia. Association
  for Computational Linguistics.

\bibitem[{Johnson et~al.(2021)Johnson, Douze, and
  J{\'{e}}gou}]{Johnson2021BillionSS}
Jeff Johnson, Matthijs Douze, and Herv{\'{e}} J{\'{e}}gou. 2021.
\newblock \href {https://doi.org/10.1109/TBDATA.2019.2921572} {Billion-scale
  similarity search with gpus}.
\newblock \emph{{IEEE} Trans. Big Data}, 7(3):535--547.

\bibitem[{Jones and Staveley(1999)}]{Jones1999PhrasierAS}
Steve Jones and Mark~S. Staveley. 1999.
\newblock \href {https://doi.org/10.1145/312624.312671} {Phrasier: {A} system
  for interactive document retrieval using keyphrases}.
\newblock In \emph{{SIGIR} '99: Proceedings of the 22nd Annual International
  {ACM} {SIGIR} Conference on Research and Development in Information
  Retrieval, August 15-19, 1999, Berkeley, CA, {USA}}, pages 160--167. {ACM}.

\bibitem[{Karpukhin et~al.(2020)Karpukhin, Oguz, Min, Lewis, Wu, Edunov, Chen,
  and Yih}]{karpukhin-etal-2020-dense}
Vladimir Karpukhin, Barlas Oguz, Sewon Min, Patrick Lewis, Ledell Wu, Sergey
  Edunov, Danqi Chen, and Wen-tau Yih. 2020.
\newblock \href {https://doi.org/10.18653/v1/2020.emnlp-main.550} {Dense
  passage retrieval for open-domain question answering}.
\newblock In \emph{Proceedings of the 2020 Conference on Empirical Methods in
  Natural Language Processing (EMNLP)}, pages 6769--6781, Online. Association
  for Computational Linguistics.

\bibitem[{Kim et~al.(2021)Kim, Jeong, Choi, and
  Hwang}]{kim-etal-2021-structure}
Jihyuk Kim, Myeongho Jeong, Seungtaek Choi, and Seung-won Hwang. 2021.
\newblock \href {https://doi.org/10.18653/v1/2021.emnlp-main.209}
  {Structure-augmented keyphrase generation}.
\newblock In \emph{Proceedings of the 2021 Conference on Empirical Methods in
  Natural Language Processing}, pages 2657--2667, Online and Punta Cana,
  Dominican Republic. Association for Computational Linguistics.

\bibitem[{Kingma and Ba(2015)}]{adam}
Diederik~P. Kingma and Jimmy Ba. 2015.
\newblock \href {http://arxiv.org/abs/1412.6980} {Adam: {A} method for
  stochastic optimization}.
\newblock In \emph{3rd International Conference on Learning Representations,
  {ICLR} 2015, San Diego, CA, USA, May 7-9, 2015, Conference Track
  Proceedings}.

\bibitem[{Lee(2013)}]{Lee2013PseudoLabelT}
Dong-Hyun Lee. 2013.
\newblock Pseudo-label : The simple and efficient semi-supervised learning
  method for deep neural networks.

\bibitem[{Lee et~al.(2019)Lee, Chang, and Toutanova}]{lee-etal-2019-latent}
Kenton Lee, Ming-Wei Chang, and Kristina Toutanova. 2019.
\newblock \href {https://doi.org/10.18653/v1/P19-1612} {Latent retrieval for
  weakly supervised open domain question answering}.
\newblock In \emph{Proceedings of the 57th Annual Meeting of the Association
  for Computational Linguistics}, pages 6086--6096, Florence, Italy.
  Association for Computational Linguistics.

\bibitem[{Lewis et~al.(2020)Lewis, Perez, Piktus, Petroni, Karpukhin, Goyal,
  K\"{u}ttler, Lewis, Yih, Rockt\"{a}schel, Riedel, and
  Kiela}]{NEURIPS2020_6b493230}
Patrick Lewis, Ethan Perez, Aleksandra Piktus, Fabio Petroni, Vladimir
  Karpukhin, Naman Goyal, Heinrich K\"{u}ttler, Mike Lewis, Wen-tau Yih, Tim
  Rockt\"{a}schel, Sebastian Riedel, and Douwe Kiela. 2020.
\newblock \href
  {https://proceedings.neurips.cc/paper/2020/file/6b493230205f780e1bc26945df7481e5-Paper.pdf}
  {Retrieval-augmented generation for knowledge-intensive nlp tasks}.
\newblock In \emph{Advances in Neural Information Processing Systems},
  volume~33, pages 9459--9474. Curran Associates, Inc.

\bibitem[{Liang et~al.(2021)Liang, Wu, Li, and
  Li}]{liang-etal-2021-unsupervised}
Xinnian Liang, Shuangzhi Wu, Mu~Li, and Zhoujun Li. 2021.
\newblock \href {https://doi.org/10.18653/v1/2021.emnlp-main.14} {Unsupervised
  keyphrase extraction by jointly modeling local and global context}.
\newblock In \emph{Proceedings of the 2021 Conference on Empirical Methods in
  Natural Language Processing}, pages 155--164, Online and Punta Cana,
  Dominican Republic. Association for Computational Linguistics.

\bibitem[{Liu et~al.(2020{\natexlab{a}})Liu, Lin, Fu, and
  Wang}]{liu2020reinforced}
Rui Liu, Zheng Lin, Peng Fu, and Weiping Wang. 2020{\natexlab{a}}.
\newblock Reinforced keyphrase generation with bert-based sentence scorer.
\newblock In \emph{2020 IEEE Intl Conf on Parallel \& Distributed Processing
  with Applications, Big Data \& Cloud Computing, Sustainable Computing \&
  Communications, Social Computing \& Networking
  (ISPA/BDCloud/SocialCom/SustainCom)}, pages 1--8. IEEE.

\bibitem[{Liu et~al.(2020{\natexlab{b}})Liu, Gu, Goyal, Li, Edunov,
  Ghazvininejad, Lewis, and Zettlemoyer}]{liu-etal-2020-multilingual-denoising}
Yinhan Liu, Jiatao Gu, Naman Goyal, Xian Li, Sergey Edunov, Marjan
  Ghazvininejad, Mike Lewis, and Luke Zettlemoyer. 2020{\natexlab{b}}.
\newblock \href {https://doi.org/10.1162/tacl_a_00343} {Multilingual denoising
  pre-training for neural machine translation}.
\newblock \emph{Transactions of the Association for Computational Linguistics},
  8:726--742.

\bibitem[{Liu et~al.(2011)Liu, Chen, Zheng, and Sun}]{liu2011gap}
Zhiyuan Liu, Xinxiong Chen, Yabin Zheng, and Maosong Sun. 2011.
\newblock Automatic keyphrase extraction by bridging vocabulary gap.
\newblock \emph{the Fifteenth Conference on Computational Natural Language
  Learning}.

\bibitem[{Luo et~al.(2021)Luo, Xu, Ye, Qiu, and
  Zhang}]{luo-etal-2021-keyphrase-generation}
Yichao Luo, Yige Xu, Jiacheng Ye, Xipeng Qiu, and Qi~Zhang. 2021.
\newblock \href {https://doi.org/10.18653/v1/2021.findings-emnlp.45} {Keyphrase
  generation with fine-grained evaluation-guided reinforcement learning}.
\newblock In \emph{Findings of the Association for Computational Linguistics:
  EMNLP 2021}, pages 497--507, Punta Cana, Dominican Republic. Association for
  Computational Linguistics.

\bibitem[{Meng et~al.(2021)Meng, Yuan, Wang, Zhao, Trischler, and
  He}]{meng-etal-2021-empirical}
Rui Meng, Xingdi Yuan, Tong Wang, Sanqiang Zhao, Adam Trischler, and Daqing He.
  2021.
\newblock \href {https://doi.org/10.18653/v1/2021.naacl-main.396} {An empirical
  study on neural keyphrase generation}.
\newblock In \emph{Proceedings of the 2021 Conference of the North American
  Chapter of the Association for Computational Linguistics: Human Language
  Technologies}, pages 4985--5007, Online. Association for Computational
  Linguistics.

\bibitem[{Meng et~al.(2017)Meng, Zhao, Han, He, Brusilovsky, and
  Chi}]{meng-etal-2017-deep}
Rui Meng, Sanqiang Zhao, Shuguang Han, Daqing He, Peter Brusilovsky, and
  Yu~Chi. 2017.
\newblock \href {https://doi.org/10.18653/v1/P17-1054} {Deep keyphrase
  generation}.
\newblock In \emph{Proceedings of the 55th Annual Meeting of the Association
  for Computational Linguistics (Volume 1: Long Papers)}, pages 582--592,
  Vancouver, Canada. Association for Computational Linguistics.

\bibitem[{Mihalcea and Tarau(2004)}]{mihalcea-tarau-2004-textrank}
Rada Mihalcea and Paul Tarau. 2004.
\newblock \href {https://aclanthology.org/W04-3252} {{T}ext{R}ank: Bringing
  order into text}.
\newblock In \emph{Proceedings of the 2004 Conference on Empirical Methods in
  Natural Language Processing}, pages 404--411, Barcelona, Spain. Association
  for Computational Linguistics.

\bibitem[{Mu et~al.(2020)Mu, Yu, Wang, Wang, Yin, Sun, Liu, Ma, Tang, and
  Zhou}]{mu2020keyphrase}
Funan Mu, Zhenting Yu, LiFeng Wang, Yequan Wang, Qingyu Yin, Yibo Sun, Liqun
  Liu, Teng Ma, Jing Tang, and Xing Zhou. 2020.
\newblock Keyphrase extraction with span-based feature representations.
\newblock \emph{arXiv preprint arXiv:2002.05407}.

\bibitem[{Park and Caragea(2020)}]{park-caragea-2020-scientific}
Seoyeon Park and Cornelia Caragea. 2020.
\newblock \href {https://doi.org/10.18653/v1/2020.coling-main.472} {Scientific
  keyphrase identification and classification by pre-trained language models
  intermediate task transfer learning}.
\newblock In \emph{Proceedings of the 28th International Conference on
  Computational Linguistics}, pages 5409--5419, Barcelona, Spain (Online).
  International Committee on Computational Linguistics.

\bibitem[{Petroni et~al.(2021)Petroni, Piktus, Fan, Lewis, Yazdani, De~Cao,
  Thorne, Jernite, Karpukhin, Maillard, Plachouras, Rockt{\"a}schel, and
  Riedel}]{petroni-etal-2021-kilt}
Fabio Petroni, Aleksandra Piktus, Angela Fan, Patrick Lewis, Majid Yazdani,
  Nicola De~Cao, James Thorne, Yacine Jernite, Vladimir Karpukhin, Jean
  Maillard, Vassilis Plachouras, Tim Rockt{\"a}schel, and Sebastian Riedel.
  2021.
\newblock \href {https://doi.org/10.18653/v1/2021.naacl-main.200} {{KILT}: a
  benchmark for knowledge intensive language tasks}.
\newblock In \emph{Proceedings of the 2021 Conference of the North American
  Chapter of the Association for Computational Linguistics: Human Language
  Technologies}, pages 2523--2544, Online. Association for Computational
  Linguistics.

\bibitem[{Pham et~al.(2021)Pham, Xie, Dai, and Le}]{Pham2021MetaPL}
Hieu Pham, Qizhe Xie, Zihang Dai, and Quoc~V. Le. 2021.
\newblock Meta pseudo labels.
\newblock \emph{2021 IEEE/CVF Conference on Computer Vision and Pattern
  Recognition (CVPR)}, pages 11552--11563.

\bibitem[{Reimers and Gurevych(2019)}]{reimers-gurevych-2019-sentence}
Nils Reimers and Iryna Gurevych. 2019.
\newblock \href {https://doi.org/10.18653/v1/D19-1410} {Sentence-{BERT}:
  Sentence embeddings using {S}iamese {BERT}-networks}.
\newblock In \emph{Proceedings of the 2019 Conference on Empirical Methods in
  Natural Language Processing and the 9th International Joint Conference on
  Natural Language Processing (EMNLP-IJCNLP)}, pages 3982--3992, Hong Kong,
  China. Association for Computational Linguistics.

\bibitem[{Shang et~al.(2018)Shang, Liu, Jiang, Ren, Voss, and
  Han}]{shang2018automated}
Jingbo Shang, Jialu Liu, Meng Jiang, Xiang Ren, Clare~R Voss, and Jiawei Han.
  2018.
\newblock Automated phrase mining from massive text corpora.
\newblock \emph{IEEE Transactions on Knowledge and Data Engineering},
  30(10):1825--1837.

\bibitem[{Sinha et~al.(2015)Sinha, Shen, Song, Ma, Eide, Hsu, and
  Wang}]{Sinha2015AnOO}
Arnab Sinha, Zhihong Shen, Yang Song, Hao Ma, Darrin Eide, Bo{-}June~Paul Hsu,
  and Kuansan Wang. 2015.
\newblock \href {https://doi.org/10.1145/2740908.2742839} {An overview of
  microsoft academic service {(MAS)} and applications}.
\newblock In \emph{Proceedings of the 24th International Conference on World
  Wide Web Companion, {WWW} 2015, Florence, Italy, May 18-22, 2015 - Companion
  Volume}, pages 243--246. {ACM}.

\bibitem[{Sterckx et~al.(2015)Sterckx, Demeester, Deleu, and
  Develder}]{Sterckx2015TopicalWI}
Lucas Sterckx, Thomas Demeester, Johannes Deleu, and Chris Develder. 2015.
\newblock \href {https://doi.org/10.1145/2740908.2742730} {Topical word
  importance for fast keyphrase extraction}.
\newblock In \emph{Proceedings of the 24th International Conference on World
  Wide Web Companion, {WWW} 2015, Florence, Italy, May 18-22, 2015 - Companion
  Volume}, pages 121--122. {ACM}.

\bibitem[{Swaminathan et~al.(2020)Swaminathan, Zhang, Mahata, Gosangi, Shah,
  and Stent}]{swaminathan2020preliminary}
Avinash Swaminathan, Haimin Zhang, Debanjan Mahata, Rakesh Gosangi, Rajiv Shah,
  and Amanda Stent. 2020.
\newblock A preliminary exploration of gans for keyphrase generation.
\newblock In \emph{Proceedings of the 2020 Conference on Empirical Methods in
  Natural Language Processing (EMNLP)}, pages 8021--8030.

\bibitem[{Wang et~al.(2016)Wang, Zhao, and Huang}]{wang2016ptr}
Minmei Wang, Bo~Zhao, and Yihua Huang. 2016.
\newblock Ptr: Phrase-based topical ranking for automatic keyphrase extraction
  in scientific publications.
\newblock \emph{23rd International Conference, ICONIP 2016}.

\bibitem[{Wang et~al.(2019)Wang, Li, Chan, King, Lyu, and Shi}]{wang2019topic}
Yue Wang, Jing Li, Hou~Pong Chan, Irwin King, Michael~R. Lyu, and Shuming Shi.
  2019.
\newblock \href {https://doi.org/10.18653/v1/P19-1240} {Topic-aware neural
  keyphrase generation for social media language}.
\newblock In \emph{Proceedings of the 57th Annual Meeting of the Association
  for Computational Linguistics}, pages 2516--2526, Florence, Italy.
  Association for Computational Linguistics.

\bibitem[{Weston et~al.(2018)Weston, Dinan, and
  Miller}]{weston-etal-2018-retrieve}
Jason Weston, Emily Dinan, and Alexander Miller. 2018.
\newblock \href {https://doi.org/10.18653/v1/W18-5713} {Retrieve and refine:
  Improved sequence generation models for dialogue}.
\newblock In \emph{Proceedings of the 2018 {EMNLP} Workshop {SCAI}: The 2nd
  International Workshop on Search-Oriented Conversational {AI}}, pages 87--92,
  Brussels, Belgium. Association for Computational Linguistics.

\bibitem[{Witten et~al.(1999{\natexlab{a}})Witten, Paynter, Frank, Gutwin, and
  Nevill-Manning}]{witten1999KEA}
Ian~H. Witten, Gordon~W. Paynter, Eibe Frank, Carl Gutwin, and Craig~G.
  Nevill-Manning. 1999{\natexlab{a}}.
\newblock Kea: Practical automatic keyphrase extraction.
\newblock In \emph{Proceedings of the Fourth ACM Conference on Digital
  Libraries}, DL '99, pages 254--255, New York, NY, USA. ACM.

\bibitem[{Witten et~al.(1999{\natexlab{b}})Witten, Paynter, Frank, Gutwin, and
  Nevill{-}Manning}]{Witten1999PracticalAK}
Ian~H. Witten, Gordon~W. Paynter, Eibe Frank, Carl Gutwin, and Craig~G.
  Nevill{-}Manning. 1999{\natexlab{b}}.
\newblock \href {https://doi.org/10.1145/313238.313437} {{KEA:} practical
  automatic keyphrase extraction}.
\newblock In \emph{Proceedings of the Fourth {ACM} conference on Digital
  Libraries, August 11-14, 1999, Berkeley, CA, {USA}}, pages 254--255. {ACM}.

\bibitem[{Wolf et~al.(2020)Wolf, Debut, Sanh, Chaumond, Delangue, Moi, Cistac,
  Rault, Louf, Funtowicz, Davison, Shleifer, von Platen, Ma, Jernite, Plu, Xu,
  Le~Scao, Gugger, Drame, Lhoest, and Rush}]{wolf-etal-2020-transformers}
Thomas Wolf, Lysandre Debut, Victor Sanh, Julien Chaumond, Clement Delangue,
  Anthony Moi, Pierric Cistac, Tim Rault, Remi Louf, Morgan Funtowicz, Joe
  Davison, Sam Shleifer, Patrick von Platen, Clara Ma, Yacine Jernite, Julien
  Plu, Canwen Xu, Teven Le~Scao, Sylvain Gugger, Mariama Drame, Quentin Lhoest,
  and Alexander Rush. 2020.
\newblock \href {https://doi.org/10.18653/v1/2020.emnlp-demos.6} {Transformers:
  State-of-the-art natural language processing}.
\newblock In \emph{Proceedings of the 2020 Conference on Empirical Methods in
  Natural Language Processing: System Demonstrations}, pages 38--45, Online.
  Association for Computational Linguistics.

\bibitem[{Ye and Wang(2018)}]{ye-wang-2018-semi}
Hai Ye and Lu~Wang. 2018.
\newblock \href {https://doi.org/10.18653/v1/D18-1447} {Semi-supervised
  learning for neural keyphrase generation}.
\newblock In \emph{Proceedings of the 2018 Conference on Empirical Methods in
  Natural Language Processing}, pages 4142--4153, Brussels, Belgium.
  Association for Computational Linguistics.

\bibitem[{Ye et~al.(2021{\natexlab{a}})Ye, Cai, Gui, and
  Zhang}]{ye2021heterogeneous}
Jiacheng Ye, Ruijian Cai, Tao Gui, and Qi~Zhang. 2021{\natexlab{a}}.
\newblock Heterogeneous graph neural networks for keyphrase generation.
\newblock \emph{arXiv preprint arXiv:2109.04703}.

\bibitem[{Ye et~al.(2021{\natexlab{b}})Ye, Gui, Luo, Xu, and
  Zhang}]{ye-etal-2021-one2set}
Jiacheng Ye, Tao Gui, Yichao Luo, Yige Xu, and Qi~Zhang. 2021{\natexlab{b}}.
\newblock \href {https://doi.org/10.18653/v1/2021.acl-long.354} {{O}ne2{S}et:
  {G}enerating diverse keyphrases as a set}.
\newblock In \emph{Proceedings of the 59th Annual Meeting of the Association
  for Computational Linguistics and the 11th International Joint Conference on
  Natural Language Processing (Volume 1: Long Papers)}, pages 4598--4608,
  Online. Association for Computational Linguistics.

\bibitem[{Zhao and Zhang(2019)}]{zhao2019incorporating}
Jing Zhao and Yuxiang Zhang. 2019.
\newblock Incorporating linguistic constraints into keyphrase generation.
\newblock In \emph{Proceedings of the 57th Annual Meeting of the Association
  for Computational Linguistics}, pages 5224--5233.

\end{thebibliography}
\bibliographystyle{acl_natbib}

\clearpage

\appendix

\section{Appendix}

\subsection{Implementation Details} 
In the cross-lingual dense passage retriever, we use ``{bert-base-multilingual-cased}'' model~\cite{wolf-etal-2020-transformers} to initialize the query and passage encoders and fine-tune it for 15 epochs with a batch size of 32.
We share the parameters between the query encoder $E_Q(\cdot)$ and the passage encoder $E_P(\cdot)$ and map English and non-English passages into the same embedding space. 
Empirical results show the encoder with parameter sharing can perform slightly better.
The positive examples are the corresponding English passages while we randomly sample 100 passages as negative examples in training.
For the retrieval-augmented keyphrase generator, we fine-tune ``{mbart-large-cc25}''~\cite{wolf-etal-2020-transformers} for 10 epochs with Adam~\cite{adam} optimizer with a learning rate of 1e-4, a batch size of 8, a warm-up rate of 50 training steps.
Similar to most Seq2Seq models, we train the mBART-based generation module by optimizing the negative log-likelihood loss of the ground-truth keyphrase sequence, and use beam search decoding with a beam size of 5 during inference.
The number of retrieved keyphrases $m$ for retrieval-augmented generation is a hyperparameter and is tuned on the development set.
We use keyphrases from $m=1$ English passages for \magdata\ dataset and $m=5$ for \ecomdata\ dataset.
During inference, we set the maximum target sequence length as 128 and set the beam decoding size as 5.
For parallel passage mining via iterative training, we continue the iterative process until the retrieval recall does not improve. The total number of iterations ($T$ in Algo.~\ref{alg:pseudo}) are 6 and 3 on \ecomdata\ and \magdata\ respectively.
The threshold $\tau$ in Line 23 for Algo.~\ref{alg:pseudo} is set as 5.

\subsection{Variants of Retrieval Targets}
There exists a misalignment between the retriever and the generator model. The retriever retrieves similar passages while the generator utilizes the associated keyphrases of these passages (not the retrieved passages) as external knowledge for generation. Therefore, a good retriever does not necessarily guarantee the good quality and usefulness of these keyphrases.

We tried two different retrieval targets which might close the misalignment.  
Given a non-English passage, we tried to either directly retrieve English keyphrases (RAMKG-P2K) or retrieve the concatenated sequences of passage-keyphrase pair (RAMKG-P2PK).
We find that (1) RAMKG-P2K that directly retrieves keyphrases has poor retrieval performance. This is because it is hard to capture the similarity between non-English passages and English keyphrases; (2)  RAMKG-P2PK has slightly worse results than only retrieval EN passages, which implies that additionally adding keyphrases in the retrieval targets does not bring any benefit.

Results are shown in Table~\ref{tab:result-variant}.
RAMKG (P2P) is our original model which retrieves English passages given a non-english passage. Results tell us that 1) directly retrieval of keyphrases have poor retrieval performance. This is because it is hard to capture the similarity between non-english passages and english keyphrases; 2)  RAMKG (P2PK)  has slightly worse results than the model   , which implies that additionally adding keyphrases in the retrieval targets does not bring any benefit.

\begin{table}[!t]
  \centering
  \resizebox{1.0\columnwidth}{!}{
    \begin{tabular}{ccccccc}
    \toprule
    \multirow{2}[2]{*}{Model} & \multicolumn{3}{c}{Retrieval Results} & \multicolumn{3}{c}{Generation Results} \\
          & \multicolumn{1}{l}{Recall@1} & \multicolumn{1}{l}{Recall@2} & \multicolumn{1}{l}{Recall@5} & P     & R     & F1 \\
    \midrule
    P2P & 26.00\% & 36.78\% & 50.05\% & 48.51 & 44.71 & 44.50 \\
    P2K & 2.86\% & 4.70\% & 9.09\% & 46.91 & 43.13 & 42.95 \\
    P2PK & 25.25\% & 35.45\% & 49.51\% & 48.50 & 44.39 & 44.34 \\
    \bottomrule
    \end{tabular}%
    }
  \caption{Results of RAMKG variants with different retrieval targets.}
  \label{tab:result-variant}%
\end{table}%

\begin{table}[!t]
  \centering
  \resizebox{0.8\columnwidth}{!}{
    \begin{tabular}{ccccc}
    \toprule
    $\tau$ & 0     & 5     & 10    & 15 \\
    \midrule
    Recall & 62.28\% & 64.01\% & 63.67\% & 62.97\% \\
    \bottomrule
    \end{tabular}%
    }
    \caption{Influence of the threshold $\tau$ on the retriever-generator self-training algorithm.}
  \label{tab:selftraining_threshold}%
\end{table}%

\subsection{Discussion on Retriever-Generator Iterative training (RGIT) Algorithm}
\paragraph{Difference between RGIT and Self-Training.}
Our approach shares some similarities with self-training \cite{Lee2013PseudoLabelT,Pham2021MetaPL} but there are some differences. 
In self-training, the teacher and student models are in the same architecture and focus on the same training objectives.
In our proposed retriever-generator iterative training, the retriever and generator are two different models and optimized by different objectives.

\paragraph{Threshold Tuning.}
In this section, we investigate the impact of the chosen threshold $\tau$ (line 24) in our proposed retriever-generator iterative training.
We tune the threshold (tau) on AcademicMKP and results are shown in Table \ref{tab:selftraining_threshold}.
Results show that tau=5 receives the best retrieval performance. Tau=0 brings more pseudo parallel passage pairs but introduces more noise. Larger tau (10/15) reduces the number of pseudo pairs, making the iterative training less effective.


\end{document}